# Expanding the Knowledge Horizon in Underwater Robot Swarms

Enzo Fioriti[1], Stefano Chiesa[1], Fabio Fratichini[1]

[1]ENEA, CR Casaccia, UTTEI-ROB Unit, Robotics Laboratory
vincenzo.fioriti, stefano.chiesa@enea.it

**Abstract**

In this paper we study the time delays affecting the diffusion of information in an underwater heterogeneous robot swarm, considering a time-sensitive environment. In many situations each member of the swarm must update its knowledge about the environment as soon as possible, thus every effort to expand the knowledge horizon is useful. Otherwise critical information may not reach nodes far from the source causing dangerous misbehaviour of the swarm. We consider two extreme situations. In the first scenario we have an unique probabilistic delay distribution. In the second scenario, each agent is subject to a different truncated gaussian distribution, meaning local conditions are significantly different from link to link. We study how several swarm topologies react to the two scenarios and how to allocate the more efficient transmission resources in order to expand the horizon. Results show that significant time savings under a gossip-like protocol are possible properly allocating the resources. Moreover, methods to determine the fastest swarm topologies and the most important nodes are suggested.

## Introduction

The robotic technology in ocean surveys, inspections, monitoring, pipe and cable tracking, has been well established in marine engineering (Leonard, 1998) with an important increase in performance in recent years (Nawaz, 2005).

Today, an AUV (Autonomous Underwater Vehicle) must be considered (Dell'Erba, 2012) as a real cost alternative to other available technologies, such as manned submersibles, remotely operated vehicles and towed instruments led by ships.

A group of underwater robots resembles closely a fish swarm, suggesting to use the properties of the biological swarm: coordinated movements, decentralized control, small interaction scale, minimal information broadcast.

The biologically inspired swarm control has advantages over the more complex but single robot: it covers a larger area, is fault tolerant, is self-aware. But it needs an inter-swarm information exchange and consequently delays during the information spread are generated (Beni, 1989).

However, communication channels are a major concern, as the acoustic underwater transmission is very slow and bandwidth limited. In the future, optical high power transmission devices will be available for a number of different approaches integrating the acoustical data channel. Although optical methods are very powerful, their performances are affected by many strongly variable parameters like temperature, depth, salinity, turbidity, the presence of dissolved substances that change the colour and the transparency in different optical bands and the amount of solar radiation, that heavily affect the signal to noise ratio. Moreover,horizontal-underwater channels are prone to multipath propagation due to refraction, reflection and scattering.

The low speed of sound is also at the origin of significant Doppler effects, divided in frequency shifts and instantaneous frequency spreading contributing to the Doppler variance of received communication signals (Otnes, 2011).

Another important issue is the energy consumption, that requires optimization techniques to prevent batteries early exhaustion.

Nevertheless, the acoustic option for underwater data transmission is still the state-of-the-art methodology, and in this paper we will refer to it. Moreover, to accomplish the above tasks it is necessary to control many submarine robots simultaneously, therefore researchers consider worthwhile to use biologically inspired models (Chiesa, 2012). In some cases, the spreading of the knowledge in large swarms may even be considered as a phase transition, whose mathematical treatment is certainly difficult as in the classical field case theory (Dell' Isola, 1987).

The swarm methodology has advantages over the more complex but single robot usage: covers a larger area, and is fault tolerant. But it needs a heavy inter-swarm information exchange and consequently delays in the information spread.

## Goals

Depending on the local underwater environment, the speed of acoustic waves in water varies around 1500 m/s. The other source of delay is the management of the acoustic channel by proper MAC (medium access control) protocols (Otnes, 2011).

In this paper, we focus on the reduction of delays when the fast propagation of short warning messages to the whole swarm is needed. We consider the existence of a small group of AUV equipped with high quality communication devices able to reduce the transmission time with their closest neighbours.

Then the point is how to allocate heterogeneous AUV into suitable swarm configurations to obtain the largest time savings, therefore extending the knowledge horizon (KH) of the swarm. To this end, we borrow from the graph theory some methods able to identify the most critical nodes with respect to the information transfer and test them numerically on several swarm configurations by means of Markov chains.

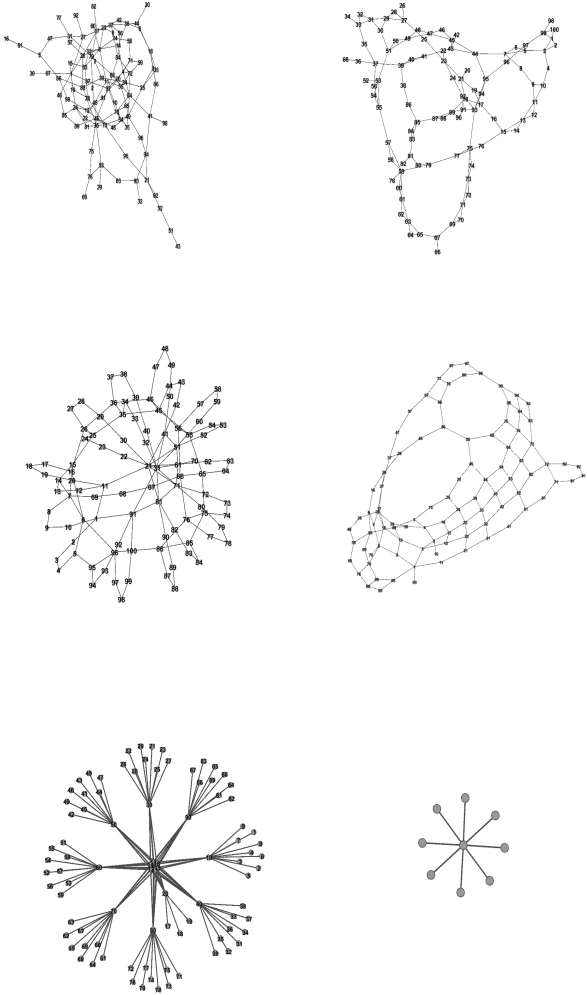

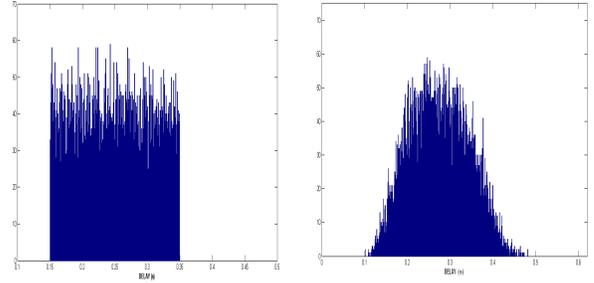

(Pompili, 2006), (Pompili, 2010) and (Burrowes, 2011).

Figure 2: Left, delays uniform distribution (scenario I); right, delays positive support Gaussian distribution (scenario II).

Figure 1: The swarm configurations. From top left: random Erdos-Renyi, Small-Word, Cluster, Grid, Galaxy. Graphs have been modified with respect to the standard topologies. Bottom right: a single star.

Technical details of transmission underwater devices are out of scope here, but can be found in a number of papers, see

## Scenarios and Data Preparation

### The Simulation Scenarios

Two scenarios are considered. In the first (scenario I), all delays are derived from an uniform probability distribution. In the second (scenario II), each link between two nodes is affected from its own Gaussian distribution with positive support, because delays are strictly positive (Mazet, 2012).

A different probability distribution for each node pair $i \rightarrow j$ has been implemented recently (Picu, 2012), but not many researchers adopt this point of view, because of the analytical intractability of the mathematical model.

Although we do not assume an explicit statistical dependence among links departing from the same node, we acknowledge the existence of a dependence (Chakrabarti, 2008) among processes that govern the delays (Xia, 2008). On the other hand, MAC protocols clearly indicate such dependences. This element is important because it rules out the usage of the Central Limit Theorem in the calculation of the overall graph delay and make necessary the heuristic approach, as opposed to the analytic approach.

The numerical values of the simulation have been elaborated according to a worst case criterion from (Pompili, 2006); they cannot be considered typical of delays encountered in the underwater environment, as its variability and non stationarity are so wide to prevent any attempt, but can represent a reasonable case study. Also the Gaussian nature of the delay distribution is to be considered an approximation of reality, since statistics are sorely lacking.

The other major source of delay, the speed of the acoustic waves in water is taken into count in the simulation model; in the scenario I the physical distances among nodes are fixed while in the scenario II may vary.

### The Configurations

The configurations assumed by the AUV (in the following called graphs or networks) considered in this paper are: the random Erdos-Renyi (ER), a graph whose nodes are connected almost randomly; the Grid, a regular disposition of nodes; the Small-World (SW), a configuration between the grid and the random graph; the Galaxy, a group of interconnected star configurations; the Cluster, a group of interconnected ring configurations.

Note that in order to have the same number of links and node for all graphs, the basic structure of the graph has been modified adding/removing links. Hence, graphs of Figure 1 *do no*t *resemble exactly* the standard topology of a small-world, of a grid or a cluster.

Moreover, the links represent the acoustic communication channel between two nodes in a logical way, therefore the geometrical form of graph in Figure 1 may change in the real environment, without loss of generality. These links are stochastic variables, meaning they may disappear randomly; their Euclidean length in the scenario I is set to 150 m, while in the scenario II can vary between 150 – 200 m, depending on a stochastic variable (if the link exists).

**Optimization Methods**

The number of the algorithms and of parameters devised to study graphs from the topological point of view is very large. Here we are interested in those who can determinate a small set of nodes equipped with particular information spreading influence. The phenomenon is analogous to the diffusion of a malware or a real virus in a computer network. It is known that some nodes are more "important" and may facilitate or prevent the spreading. In most cases, it is not a trivial task to find the influential nodes in large graphs, but fortunately AUV swarms are reduced to less than a few hundred vehicles. Nevertheless, useful insights may be gained even with just one hundred nodes, as in the present case.

The methodologies we use are: AV11 (Arbore, 2013), degree centrality, betweenness centrality and the random choice (in order to check results against the banal predictor). Other parameters are available, but AV11, degree, betweenness are probably the more relevant to our investigation. The degree centrality is the simplest algorithm, since it suffices to count the number of links connected to a node. It is also intuitive that an high degree designate an "influent" node (called hub).

Betweenness centrality (BC) is the total number of shortest paths between every possible pair of nodes that pass through a given node. Betweennes looks for vertices connecting separated subgraphs (Newman, 2010), therefore high BC nodes may be understood as a sort of bridges.

AV11 selects a subset of *k* nodes all at once, according to spectral combinatorial methods. The selected subset may be optimal or suboptimal with respect to the brute-force method (Arbore, 2013). usually spectral methods are able to analyze the dynamical behavior of graphs from the static appearance of the topology.

We ask the algorithms to identify ten nodes: for example, in the case of the Cluster, the degree centrality may select the upper row of nodes below and the AV11 the second row:

**31**, **21**, **6**, **61**, 71, **81**, 41, 91,  1, 51
**31**, **21**, **6**, **61**, 56, **81**, 76, 36, 45, 15

(in bold the common choices).

Once the set is identified, a high quality transmission AUV vehicle is assigned to the nodes. We simulate this operation assigning smaller delay capability to the links departing from the "optimized" node, i.e. optimally allocating the resources.

**Protocol**

The protocol we consider here is a gossip – like (sometimes called "epidemic spreading") protocol, i.e. each node-agent in the swarm broadcasts the message to the one-hop neighbour. The diffusion process is started by one node and must reach every node of the swarm within a finite time (the knowledge horizon, briefly KH). In this paper we follow a small number of specifications, namely:

- the transmitting node contacts the receiver following the numeration priority rule (node j is contacted before node j+1);
- agents know only their neighbours;
- the receiving node has no previous knowledge of the message to be delivered;
- the link between two nodes i → j is fixed, but the on/off status depends on a stochastic variable;
- if a link is off, no attempt to transmit is made;
- all particular delays due to MAC parameters are included into the stochastic delay characterizing the i → j link;
- apart from re-transmitting the message (eventually), no elaboration takes place inside the agent on the message data;
- the QoS (quality of service) of the communication link is completely described by the delay.

This protocol is quite general to encompass many actual protocols. It is configured to serve as warning signalling tool, therefore the message to be transmitted is very short (256B).

## Simulation Results

Numerical simulations have been conducted till stable results have been obtained. In Table 1 are shown the delays in the two scenarios. Absolute times should not be regarded as indicative of a general behaviour, because of the too wide variation of the conditions in the real marine environment.

The important point here is the relative time difference among the topologies in each scenario. Table 1 shows that the variances are small and similar, hence the numerical experiments are reliable.

In Table 2 the spectral analysis of the five topologies is presented. We rank them according to three well-known parameters of the spectral graph theory: the maximum eigenvalue of the adjacency matrix of the graph, $\lambda_n$, the spectral gap $\lambda_{n-1} - \lambda_n$ of the laplacian matrix, and the algebraic connectivity $\lambda_2$ (the first non zero eigenvalue of the laplacian matrix), given the ascending ordering:

$\lambda_1 < \lambda_2 < ... \lambda_{n-1} < \lambda_n$

| Graph | Average delay I | Delay Var. I | Average delay II | Delay Var. II |
|---|---|---|---|---|
| ER | 23.29 | 2.22 | 26.68 | 3.15 |
| SW | 23.33 | 2.37 | 26.64 | 2.85 |
| Galaxy | 23.32 | 2.16 | 25.70 | 3.00 |
| Grid | 23.35 | 2.25 | 25.77 | 2.71 |
| Cluster | 23.34 | 1.87 | 25.75 | 2.86 |

Table 1 First two columns: overall delays and variances for different topologies with unique distribution scenario I (without optimization). Last two columns: overall delays and variances for different topologies (without optimization) in the multiple distributions scenario II. The time unit is the second.

A large value of these parameters describe the connectivity of the graph, the lack or presence of isolated components, the lack or presence of bottlenecks, see (Restrepo, 2006) and (Van Mieghem, 2011). Note that information in well connected graph travel easily, thus the probability that a node may remain disconnected is lower.

The maximum eigenvalue of the adjacency matrix and the spectral gap are in complete accordance with the largest percentage reduction in the reduction of the overall delays (described in Table 3 and 4), while the algebraic connectivity differs only for the misclassification of the Grid and the Small-World topologies.

Therefore, the spectral analysis is able to predict those swarm topologies prone to realize a major delay improvement allocating the best transmission resources in the nodes suggested by one of the topological algorithms.

Now, there is a trade-off on the largest eigenvalue of the adjacency matrix in the consensus problems of Olfati-Saber. According to (Olfati-Saber, 2004 and 2007) the stability of a fixed configuration is guaranteed iff:

$$\tau \leq \pi / 2\lambda_{max} \qquad (1)$$

where $\tau$ is the uniform delay experienced by the consensus distributed computations and $\lambda_{max}$ is the maximum eigenvalue of the adjacency matrix.

Similar constraints may be set for non uniform switching topologies. The delay $\tau$ depends on a number of factors: CPU power, data transmission bandwidth, MAC protocols, the number of AUV, the inter-symbol interference. A small $\lambda_{max}$ means the network is stable against a large delay.

But, at the same time, we need to have a large $\lambda_{max}$ to take advantage of a high connectivity. The spectral analysis of Table 2 therefore helps in determining the best configuration with respect to delays and connectivity.

Table 3 an 4 show the time saving gained using the four methodologies of optimization. Table 3 concerns the scenario I, where a unique uniform distribution provides all the delays between two communicating AUV (the nodes of the graph). Only the relative percentage values should be considered, since the absolute values are strictly related to the particular framework examined. It is clear that the Galaxy (group of connected stars) configuration allows the largest delay reduction in both scenarios; in fact, the star topology (Figure 1, bottom right) is well-known to guarantee good performance and reliability, although expensive.

The classification of the configurations according to the best delay after the optimization considering both scenarios is as follows: Galaxy, Cluster, Erdos-Renyi, Grid, Small-World.

The Galaxy configuration of the AUV acoustic communications is the best by far. It is worth mentioning, anyway, that a star configuration relays heavily on the centre-star to pass and sort messages, thus dissipates too much energy for the marine environment.

A solution is to alternate the centre-star role among the AUV of the star (Snels, 2013), possibly using a frequency coding to avoid signal overlapping, if enough high quality AUV are available. Otherwise the random Erdos-Renyi graph guarantees a good performance in a number of underwater tasks.

| Graph | Max Adj eigenval | Spectral gap | Algebraic connectivity |
|---|---|---|---|
| Galaxy | 7.9426 | 4.56030 | 0.16927 |
| Cluster | 5.1077 | 1.79950 | 0.12621 |
| ER | 3.9286 | 0.77056 | 0.10797 |
| Grid | 3.4687 | 0.23352 | 0.01066 |
| SW | 3.3473 | 0.15720 | 0.05807 |

Table 2 Spectral characteristics of the graphs examined. All graphs are of 100 nodes and 140 edges, 2.8 average degree. Simulations were at least of 1000 runs. Topologies have been modified adding/removing edges to ensure the same number of edges and nodes and the same overall broadcasting delay within a ±0.2s interval. The length of the message to be transmitted is 256 bit in both scenarios.

| Grap | AV11 | Degree | BC | Random |
|---|---|---|---|---|
| ER | (15.90) +0.78% | **(15.76) -32%** | (16.83) +6.7% | (20.10) +27.5% |
| SW | (18.89) +9.3% | **(17.28) -25.9%** | (18.32) +6% | ((20.37) +17.8% |
| Galaxy | (9.28) +143% | **(3.81) -83%** | (5.00) +31% | (18.88) +395% |
| Grid | **(17.38) -25.5%** | (18.75) +7.8% | (20.07) +15% | (19.52) +9.16% |
| Cluster | **(15.27) -34.5%** | (16.38) +7% | (16.21) +6.15% | (19.90) +30% |

Table 3 Overall delay for different topologies in the single distribution scenario (I scenario), with optimization. Values within a ±0.2s confidence interval. The optimization techniques are: the AV11 algorithm, the degree centrality, the betweenness centrality and the complete random choice of nodes. Delays are expressed in seconds. Inside the brackets the absolute values. Percentages in bold are the best time saving percentages with respect to the overall delays of Table 1 for, a given graph.

We focus on the performance of the optimization algorithms (degree, AV11, BC and random choice) in selecting the most suitable nodes.

Not surprisingly, in both scenarios the degree is the optimal choice in the case of ER, SW, Galaxy graph, but AV11 scores -7% with respect to degree for the Cluster and the Grid graph.

Then the simplest of the optimization algorithms (degree centrality) is not always the most successful, as an intuitive reasoning could expect. In fact, the AV11 exceed the degree in the Grid and in the Cluster cases while in the Erdos – Renyi case differences are minimal.

When the configuration is as simple as the Galaxy (a group of stars), certainly the best choice is the degree centrality, but when the graph is more structured, more sophisticated algorithms such as AV11 are to be considered.

This is an useful finding, as often the static parameters (degree, betweenness, closeness etc.) are not able to identify correctly the influential nodes. In fact, standard parameters are correlated (Valente, 2008), therefore their resulting measurements are somewhat redundant preventing a deeper insight in the graph topology.

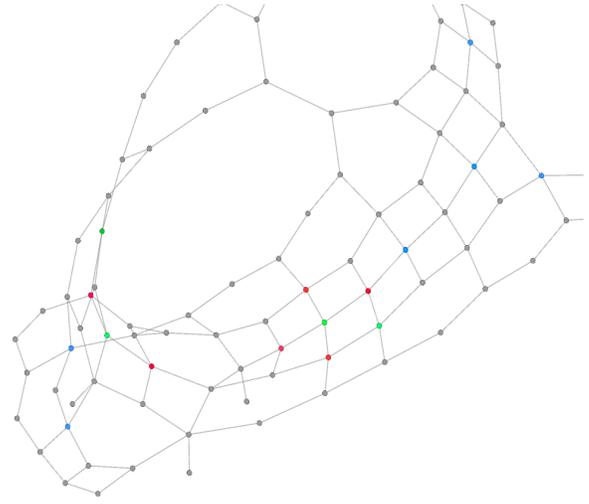

Figure 3 Comparison between degree centrality and AV11 for the case of the modified Grid graph. Green node have been selected by both algorithms, red nodes only by degree and blue node only by AV11. An intuitive ranking of importance is not straightforward in this circumstance.

## Theoretical Considerations

Now we give some theoretical hints about the effective solution of the broadcasting problem.

Chandra (Chandra, 1996) has shown how consensus and broadcasting are reducible to each other also in asynchronous networks with failure detectors. Broadcast allows processes to distribute messages, so that they agree on the set of messages they deliver and the order of message deliveries.

The standard consensus form is:

$$x_i' = \sum_j a_{ij}(x_j - x_i) \quad , \quad j = 1, ... N \quad (2)$$

and its delayed version:

$$x_i' = \sum_j a_{ij}(x_j(t - \tau) - x_i) \quad , \quad j = 1, ... N \quad (3)$$

where $a_{ij}$ are the entries of the adjacency matrix **A**.

(2) and similar expressions are utilized in the swarm control to coordinate the states of the robots on a common position/velocity agreement resilient to disturbs (Olfati Saber, 2007).

Since in the protocol section we have not required synchrony, we can use the equivalence between the consensus problem and the broadcast problem (consisting in delivering a set of messages in the correct sequence to every agent of the network) to state a sufficient condition (Lu, 2011) for the successful broadcast of a set of subsequent short messages.

Lu shows that asynchronous consensus with stochastic delays can be obtained if during motion the swarm's graph has always had a spanning tree. The same result applies equally well to our broadcasting gossip-like protocol. In a few words, this means that each node is reachable now if it was connected in the past, even in presence of (finite) delays.

Recently (Atay, 2013, in press) a necessary and sufficient condition has been obtained for the discrete time, but only for fixed, non-switching links.

| Graph | AV11 | Degree | BC | Random |
|---|---|---|---|---|
| ER | (17.33) +0.5% | **(17.24) -35.4%** | (18.54) +7.5% | (22.19) +28.7% |
| SW | (20.75) +11% | **(18.68) -29%** | (20.12) +7 | (22.39) +19 |
| Galaxy | (10.00) +161% | **(3.83) -85%** | (5.22) +36% | (20.98) +447% |
| Grid | **(19.14) -25%** | (20.75) +8.5% | (22.44) +17 | (21.16) +11.5% |
| Cluster | **(17.56) -31.8%** | (18.92) +7.75% | (18.51) +5.4% | (22.36) +27% |

Table 4      Overall delays for different topologies with optimization in the multiple distributions scenario (II scenario). The optimization techniques are: the AV11 algorithm, the degree centrality, the betweenness centrality and the complete random choice of nodes. Delays are expressed in seconds, inside the brackets are the absolute values. Percentages in bold are the best time saving percentages with respect to the overall delays of Table 1, for a given graph.

## Conclusions

We have examined the delays produced during the information diffusion in an underwater autonomous robot swarm in order to extend its knowledge horizon properly selecting the swarm configuration and the allocation of high quality transmission devices inside the configuration. Since the independence condition of variables in the analytical treatment of the delay model cannot be assured generally, numerical simulations have been conducted.

Two scenarios have been simulated. Scenario I considers a unique uniform distribution of delays with fixed inter vehicle distances; scenario II considers random distances and each link producing a delay, according to a positive support Gaussian distribution (as a delay is an inherently positive quantity). In both cases the links may or may not exist, depending on a uniform distribution probability.

Five swarm configurations (Galaxy, Cluster, Erdos-Renyi, Grid, Small-World) and four allocation methodologies (AV11, degree centrality, betweennneess, random choice) have been tested to find the largest time savings.

Results show that the degree centrality applied to the Galaxy configuration allows the largest delay reduction in both scenarios, as expected, but only when configuration are dominated by hubs. When the graphs are more structured, spectral allocation algorithms as AV11 are a better choice (Table 3 and 4).

Another useful finding is given (Table 2): the maximum adjacency eigenvalue and the spectral gap reveal the graph capabilities to decrease the delays when optimized.

**Acknowledgments**. This work has been supported by the HARNESS Project, funded by the Italian Institute of Technology (IIT).